\documentclass[lettersize,journal,twoside]{IEEEtran}
\usepackage{amssymb}
\usepackage{booktabs} % For better table rules
\usepackage[table,xcdraw]{xcolor} % For row colors
\usepackage{caption}
\usepackage{setspace} % For line spacing adjustments
\usepackage{xspace}
\usepackage{todonotes}
\usepackage[ruled,vlined]{algorithm2e} % for algorithms
\usepackage{graphicx}
\usepackage{algpseudocode}
\usepackage{siunitx}
\usepackage{svg}
\usepackage{float}
\usepackage{subcaption}
\usepackage{inconsolata}
\usepackage{bbding}
\usepackage{amsfonts}
\usepackage{wrapfig}
\usepackage{amsmath,bm}
\usepackage{hyperref}
\usepackage{tcolorbox}
\usepackage{tabularx}
\usepackage{array}
\usepackage[normalem]{ulem}


\definecolor{Gray}{gray}{0.5}
\definecolor{darkGray}{gray}{0.2}
\definecolor{newGray}{rgb}{0.827, 0.827, 0.827}
\definecolor{NewBlue}{rgb}{0.95, 0.95, 1.0}
\definecolor{Maroon}{rgb}{0.502, 0.0, 0.0}
\definecolor{Navy}{rgb}{0.0, 0.0, 0.502}
\definecolor{Olive}{rgb}{0.502, 0.502, 0.0}
\definecolor{Teal}{rgb}{0.0, 0.502, 0.502}
\definecolor{Lavender}{rgb}{0.902, 0.902, 0.98}
\definecolor{darkBlue}{rgb}{1.0, 0.95, 1.0}

\hypersetup{
	colorlinks=true,
	linkcolor=magenta,
	filecolor=magenta,      
	urlcolor=magenta,
	citecolor=orange,
}

\DeclareSIUnit{\px}{px}

\newcommand{\algabrvname}{DeCo\xspace}
\IEEEoverridecommandlockouts  
\title{\LARGE \bf{DeCo: Task Decomposition and Skill Composition for Zero-Shot Generalization in Long-Horizon 3D Manipulation}}

\author{Zixuan Chen$^{1}$, Junhui Yin$^{1}$, Yangtao Chen$^{1}$, Jing Huo$^{1\dagger}$, 
\\Pinzhuo Tian$^{2}$, Jieqi Shi$^{1}$, Yiwen Hou$^{3}$,  Yinchuan Li$^{4}$, Yang Gao$^{1}$
\thanks{
Manuscript received: August, 8, 2025; Revised November, 29, 2025; Accepted January, 25, 2026. 
This paper was recommended for publication by Editor Jlia Borrs Sol upon evaluation of the Associate Editor and Reviewers comments.
This work is supported in part by New Generation Artificial Intelligence-National Science and Technology Major Project (2025ZD0122904), National Natural Science Foundation of China (62192783, 62276128, 62506153), Jiangsu Science and Technology Major Project (BG2025035), the Fundamental Research Funds for the Central Universities (KG202514), the Collaborative Innovation Center of Novel Software Technology and Industrialization and the Postgraduate Research \& Practice Innovation Program of Jiangsu Province (KYCX25\_0317). \textit{(Corresponding author: Jing Huo.)}}
\thanks{$^{1}$Zixuan Chen, Junhui Yin, Yangtao Chen, Jing Huo, Jieqi Shi, Yang Gao are with the State Key Laboratory for
Novel Software Technology, Nanjing University, Nanjing 210093, China. (e-mail: chenzx@nju.edu.cn; yinjunhui@smail.nju.edu.cn; yangtaochen@smail.nju.edu.cn; huojing@nju.edu.cn; jayceesjq@gmail.com; gaoy@nju.edu.cn)
}
\thanks{$^{2}$Pinzhuo Tian is with the School of Electrical and Electronic Engineering, Nanyang Technological University, Singapore 639798. (e-mail: pinzhuo.tian@ntu.edu.sg)
}
\thanks{$^{3}$Yiwen Hou is with the School of Computing, National University of Singapore, Singapore 119077. (e-mail: yiwenhou@u.nus.edu)
}
\thanks{$^{4}$Yinchuan Li is with the Huawei Noah's Ark Lab (AI Lab), China. (e-mail: yinchuan.li.cn@gmail.com)
}
\thanks{Digital Object Identifier (DOI): see top of this page.}
}

\begin{document}

\maketitle

\begin{abstract}
    Generalizing language-conditioned multi-task imitation learning (IL) models to novel long-horizon 3D manipulation tasks is challenging. To address this, we propose \textbf{DeCo} (\textit{Task \textbf{De}composition and Skill \textbf{Co}mposition}), a model-agnostic framework that enhances zero-shot generalization to compositional long-horizon manipulation tasks.
    DeCo decomposes IL demonstrations into modular atomic tasks based on gripper-object interactions, creating a dataset that enables models to learn reusable skills. At inference, DeCo uses a vision-language model (VLM) to parse high-level instructions, retrieve relevant skills, and dynamically schedule their execution. A spatially-aware skill-chaining module ensures smooth, collision-free transitions between skills.
    We introduce \texttt{DeCoBench}, a benchmark designed to evaluate compositional generalization in long-horizon manipulation tasks.
    DeCo improves the success rate of three IL models—RVT-2, 3DDA, and ARP—by \textbf{66.67\%, 21.53\%, and 57.92\%}, respectively, on 12 novel tasks. In real-world experiments, the DeCo-enhanced model, trained on only 6 atomic tasks, completes 9 novel tasks in zero-shot, with a \textbf{53.33\%} improvement over the baseline model.
    Project website: \url{https://deco226.github.io}.
\end{abstract}


%===============================================================================

\section{Introduction}
\begin{figure}[t!]
    \centering
    \includegraphics[width=0.92\linewidth]{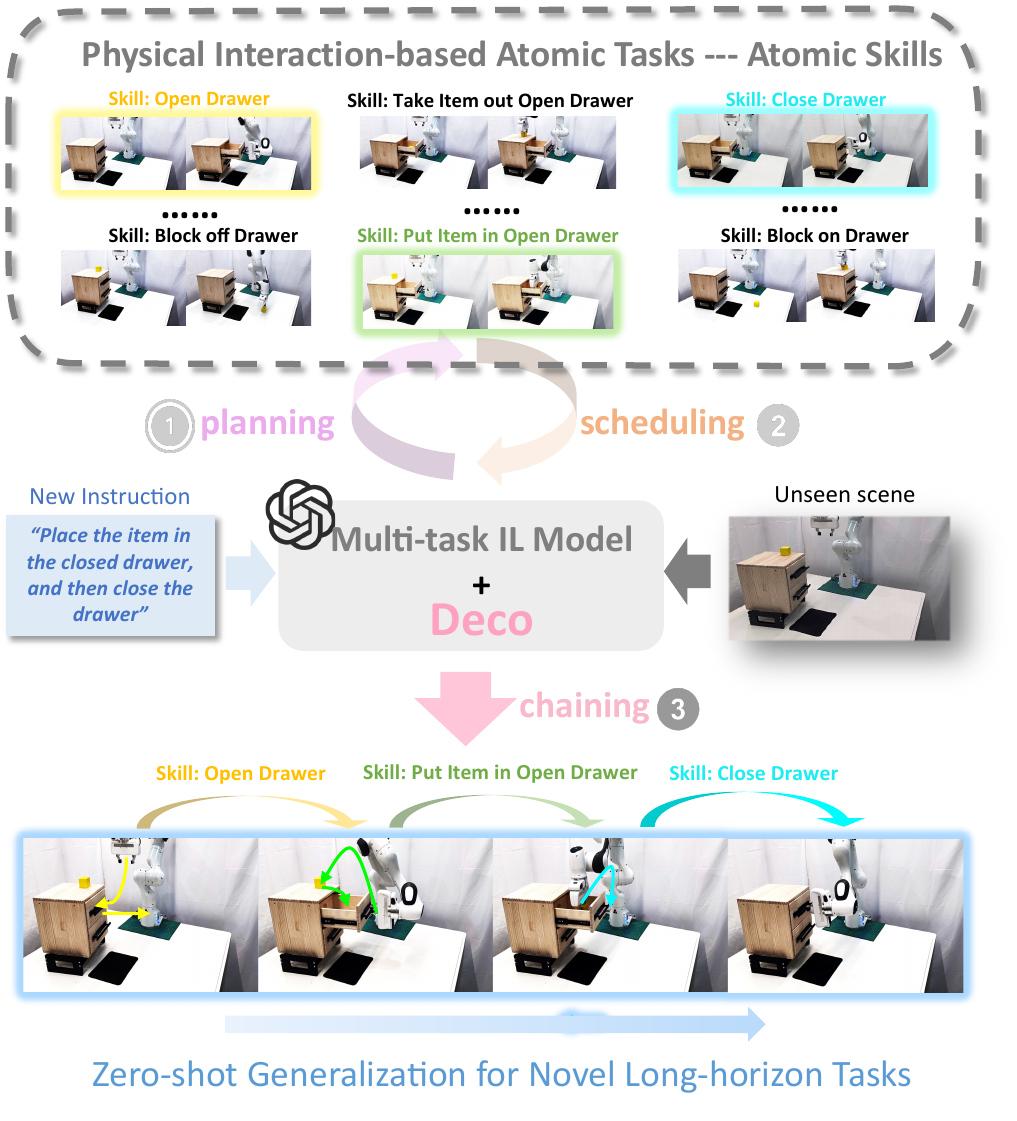}
    % \vspace{-3mm}
    \caption{
    \small We propose DeCo, a model-agnostic framework for zero-shot generalization in compositional long-horizon 3D manipulation.
    }
    \label{fig:teaser}
    % \vspace{-5mm}
\end{figure}
In recent years, imitation learning (IL) has emerged as a mainstream way for robotic manipulation. By leveraging visual demonstrations and language instructions, IL trains language-conditioned multi-task control policies, enabling robots to acquire diverse skills and perform complex tasks in unstructured 3D environments. However, current multi-task IL models still suffer from limited generalization~\cite{shridhar2023perceiver,goyal2023rvt,goyal2024rvt2,jiang2025rethinkingbimanualroboticmanipulation,zhang2025autoregressive}, particularly when facing novel long-horizon 3D manipulation tasks~\cite{garcia2024towards}---even when such tasks are merely sequential compositions of previously learned skills.
For instance, a model may have learned to follow individual instructions such as ``\textit{open drawer}", ``\textit{put block in opened drawer}", and ``\textit{close drawer}", yet still fail to execute the novel instruction ``\textit{put block into the closed drawer and then close drawer}". This failure stems from the model’s inability to decompose novel tasks and to retrieve, schedule, and perform the correct composition of its learned skills---failing to recognize that the task can be completed by sequentially executing three known skills: \textit{opening the drawer, placing the block, and then closing the drawer}.
Such limitations in task decomposition and skill composition severely undermine the real-world applicability and scalability of current multi-task IL models.
Although vision-language models (VLMs) have been used to generate subtasks for long-horizon tasks via instruction plans~\cite{myerspolicy,curtistrust,garcia2024towards}, executable code~\cite{liang2023code,chen2025robohorizon}, spatial keypoints~\cite{huang2024rekep,hao2025disco}, or affordance maps~\cite{huang2023voxposer,chen2024gravmad}, they often fail to align high-level semantic plans with low-level execution. Low-level tasks are typically limited to simple motion planning or pretrained skills, and the semantic decomposition does not directly map to the physical skill space. This gap limits the effective composition of low-level skills, ultimately hindering zero-shot performance on long-horizon 3D manipulation tasks~\cite{chen2024scar,chen2023sequential,tziafas2024lifelong}.

In this paper, we address the following core question:
\textit{How can long-horizon 3D manipulation tasks be decomposed into learned skills such that multi-task imitation learning (IL) models can interpret their structure, plan accordingly, and successfully complete novel, compositional tasks without additional training?}
To this end, we propose \textbf{DeCo} (\textit{Task \textbf{De}composition and Skill \textbf{Co}mpositon}), a model-agnostic framework compatible with a wide range of multi-task IL models. DeCo enhances the zero-shot generalization of multi-task IL models to novel, compositional long-horizon 3D manipulation tasks—tasks that are unseen during training but can be solved by composing previously learned skills through visual and semantic reasoning, as illustrated in \autoref{fig:teaser}. Specifically, DeCo enables IL models to decompose novel tasks into reusable atomic skills, flexibly schedule them, and execute skill sequences without additional training.

DeCo consists of three key components.
First, inspired by how humans decompose long-horizon tasks through hand–object interactions, DeCo proposes a new perspective on training datasets for multi-task IL, building upon prior methods of sub-task discovery in manipulation~\cite{chen2024gravmad,saito2023structured}. It preprocesses original IL demonstrations by analyzing the physical interactions between the gripper and objects, decomposing them into a set of modular and reusable atomic tasks. Each task is paired with a natural language instruction and a goal pose, forming a training dataset of atomic tasks for training multi-task IL models to acquire diverse skills.
Second, during testing, DeCo uses vision–language models (VLMs) to parse novel language instructions and visual inputs, retrieve relevant atomic instructions from the training dataset, and generate an execution plan. The multi-task IL model sequentially executes the skills, while DeCo monitors task progress via gripper interactions, enabling dynamic scheduling and flexible skill composition.
Finally, to ensure smooth transitions between skills, DeCo builds a spatially aware cost map for the scene to calculate collision-free chaining poses, guiding the robot between sequential skills and ensuring motion continuity and safety.
Extensive evaluations utilize \texttt{DeCoBench}, a benchmark built upon RLBench~\cite{james2020rlbench} to systematically evaluate zero-shot compositional generalization. Three representative models—RVT-2~\cite{goyal2024rvt2}, 3DDA~\cite{ke20243d}, and ARP~\cite{zhang2025autoregressive}—are integrated with DeCo, showing significantly improved generalization performance. In real-world settings, 6 atomic training tasks enable the zero-shot execution of 9 novel long-horizon tasks, validating practical applicability.

In summary, our main contributions are as follows:
(1) The proposal of \textbf{DeCo}, a model-agnostic framework designed to equip multi-task IL models with zero-shot generalization capabilities for novel yet compositional long-horizon 3D manipulation tasks.
(2) The development of \texttt{DeCoBench} to systematically evaluate compositional generalization, complemented by extensive real-world validation. Results demonstrate that DeCo enhances the performance of representative multi-task IL models and achieves robust zero-shot generalization on novel long-horizon tasks.

%===============================================================================
\section{Related Work}

This section reviews recent advancements in learning manipulation policies from demonstrations and strategies for long-horizon manipulation, highlighting the persistent challenges in achieving zero-shot compositional generalization.

\textbf{Learning Manipulation Policies from Demonstrations}
Learning manipulation policies from offline visual demonstrations has garnered significant attention, fueled by advances in visual perception~\cite{dosovitskiy2020image,JunchiICRA2022}. Early 2D-based approaches~\cite{chi2023diffusion,zeng2021transporter,brohan2022rt,shridhar2022cliport,chen2024scar,chen2023casil,chen2023tild} have demonstrated success in simple pick-and-place tasks, benefiting from fast training, low hardware requirements, and modest computational demands. However, their reliance on pretrained image encoders and limited spatial understanding makes them less effective for tasks requiring high-precision and robust 3D interactions.
To address this, works such as C2F-ARM~\cite{james2022coarse} and PerAct~\cite{shridhar2023perceiver} extend learning to 6-DoF actions in 3D environments, but they still require training separate task-specific policies. More recent efforts~\cite{goyal2023rvt,guhur2023instruction,gervet2023act3d,xian2023chaineddiffuser,chen2024gravmad,goyal2024rvt2,ke20243d,zhang2025autoregressive,garcia2024towards} aim to develop unified multi-task imitation learning (IL) models that can perform diverse tasks from heterogeneous demonstrations. This shift is crucial for building general-purpose robotic agents. However, most of these models are limited to tasks observed during training, and particularly struggle to generalize to novel long-horizon scenarios, which hinders their deployment in real-world applications~\cite{garcia2024towards}.
To address this, we propose a model-agnostic framework compatible with multi-task IL models for zero-shot generalization to novel long-horizon tasks.

\begin{figure*}[t!]
    \vspace{5pt}
    \centering
     \includegraphics[width=0.9\textwidth]{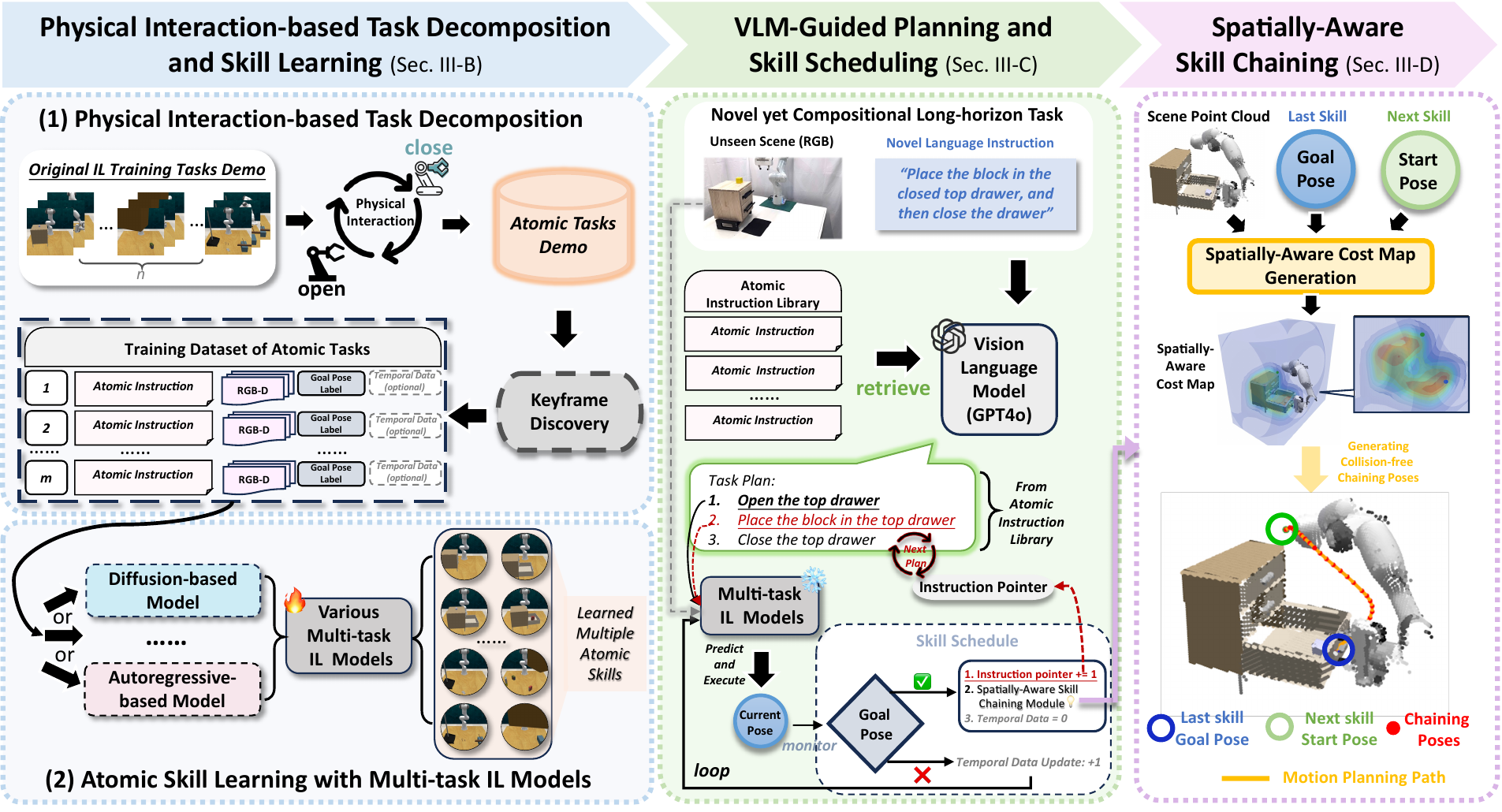}
     \vspace{-0.5mm}
    \caption{\small \textbf{The overview of DeCo framework.}
    DeCo includes three key components: 1) Physical Interaction-based Task Decomposition and Skill Learning. 2) VLM-Guided Planning and Skill Scheduling. 3) Spatially-Aware Skill Chaining.}
    \label{fig:overview}
    % \vspace{-5mm}
\end{figure*}

\textbf{Methods for Long-Horizon Manipulation}
A common strategy for long-horizon manipulation is to decompose complex tasks into sequential subtasks using predefined action primitives (e.g., grasp, place, pull)~\cite{gao2024prime,mishra2023generative,agia2023stap} or environment-specific cues~\cite{chen2024scar,hou2024effective,zentner2024conditionally,zhang2024universal,zhao2023erra,cheng2023league,chen2025manilong}. While effective in structured settings, these approaches lack compositional flexibility and generalization, making them fragile to goal shifts and environmental changes, and limiting scalability in multi-task IL.
Meanwhile, VLMs have shown promise in high-level planning by decomposing instructions into natural language, executable code, spatial keypoints, or affordance maps~\cite{myerspolicy, curtistrust, liang2023code, chen2025robohorizon, huang2023voxposer, chen2024gravmad}. However, bridging high-level semantic plans with flexible low-level execution remains challenging. For example, Points2Plans~\cite{huang2025points2plans} grounds LLM plans via relational dynamics, but still relies on parameterized action primitives rather than adaptive policies. As a result, low-level behaviors remain constrained to predefined skills, and semantic decomposition often fails to align with the physical skill space, limiting composition and generalization in long-horizon 3D manipulation~\cite{chen2024scar,chen2023sequential,tziafas2024lifelong}.
To address these limitations, we propose a modular and reusable atomic task construction that enables consistent decomposition across diverse scenarios and compatibility with multi-task IL models. We further introduce a spatially-aware skill chaining module with collision avoidance. Combined with VLM-guided planning, our framework improves generalization and robustness in compositional long-horizon manipulation.
\section{Method}

\begin{figure}[h]
    \centering
    \includegraphics[width=\linewidth]{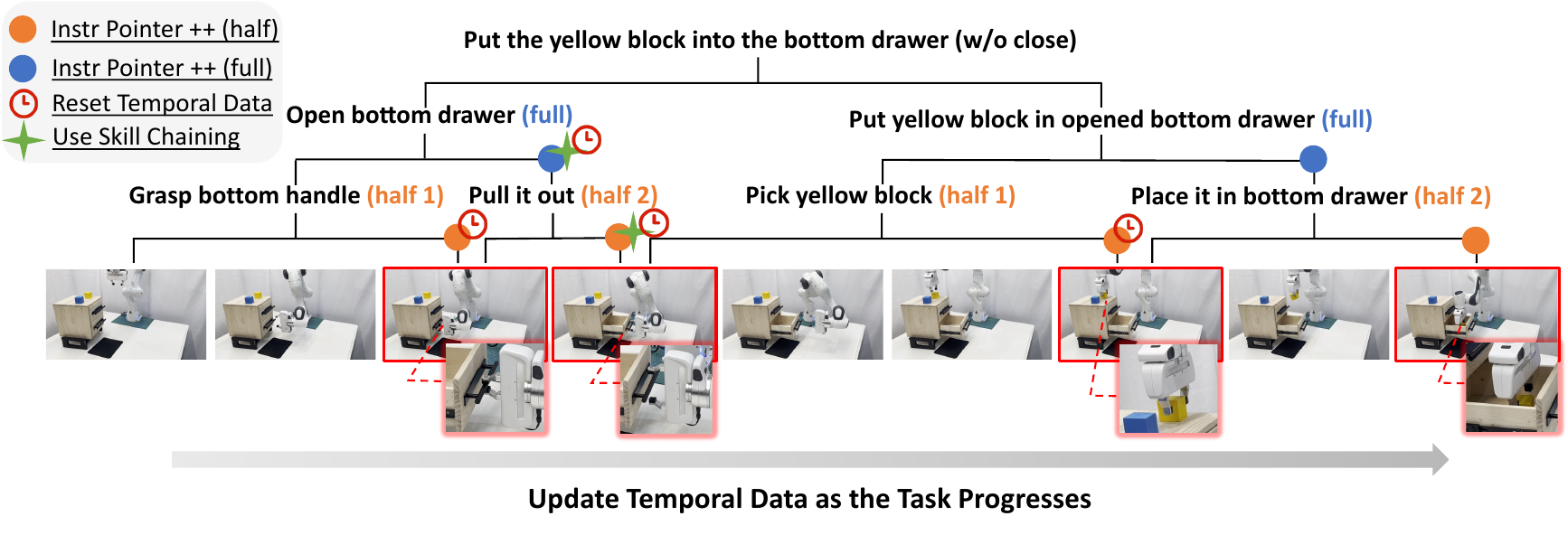}
    \caption{
    \textcolor{black}{\small \textbf{Half vs. Full Interactions.} 
    }
    }
    \label{fig:half_full_diff}
\end{figure}

\subsection{Problem Formulation}\label{method:formula}
We describe the end-effector pose with the position vector and orientation unit quaternion of the gripper, which fully characterize its spatial state and serve as the key action parameter in manipulation tasks.  
We define the \textit{physical interaction} as the contact event between a robotic gripper and an object, identified by changes in the gripper's openness. A single change in the gripper (open to closed or vice versa) is a \textit{cycle}. A full interaction, denoted as $p^{\texttt{full}}$, consists of two cycles: open $\rightarrow$ closed $\rightarrow$ open. A half interaction, $p^{\texttt{half}}$, represents a single change from open to closed or vice versa:  
$p^{\texttt{full}} = p^{\texttt{half}}_{o \to c} \rightarrow p^{\texttt{half}}_{c \to o},$  
where $p^{\texttt{half}}_{o \to c}$ and $p^{\texttt{half}}_{c \to o}$ represent the two sub-phases of the gripper transition. A visual illustration of full and half interactions is shown in \autoref{fig:half_full_diff}.  
We assume access to a training task set $\mathcal{T}^o = \{T^o_1, T^o_2, \dots, T^o_n\}$ (subscript $o$ denotes \textit{original} tasks), each paired with a natural language instruction $\ell^o_i$. However, the physical interaction phases within each $T^o_i$ are often inconsistent. By decomposing tasks using predefined interaction boundaries, we construct an atomic task set $\mathcal{T}^{a} = \{T^a_1, T^a_2, \dots, T^a_m\}$ (subscript $a$ denotes \textit{atomic} tasks, typically $m > n$) and a corresponding instruction library $\mathcal{L}^{a} = \{\ell^a_1, \ell^a_2, \dots, \ell^a_m\}$. Each atomic task contains a consistent interaction cycle, either $p^{\texttt{full}}$ or $p^{\texttt{half}}$.  
We frame 3D manipulation as keyframe prediction~\cite{james2022q,james2022coarse,goyal2023rvt,shridhar2023perceiver,chen2024gravmad}.A language-conditioned multi-task IL model $\mathcal{M}$ takes as input the observation $o_t$ (RGB-D) and instruction $\ell$, and predicts a 6-DoF end-effector pose and a 1-DoF gripper state~\cite{shridhar2023perceiver,goyal2023rvt}.
Trained on $\mathcal{T}^{a}$, the model $\mathcal{M}$ learns a policy $\pi$ to solve any $T^a_i \in \mathcal{T}^{a}$. Our objective is for the enhanced model $\mathcal{M} + \text{DeCo}$ to generalize zero-shot to a \textbf{novel compositional long-horizon task $T^{\text{new}}$}, decomposable into atomic steps:  
$T^{\text{new}} = T^a_x \rightarrow T^a_y \rightarrow \dots \rightarrow T^a_z.$  
At inference, $\mathcal{M} + \mathrm{DeCo}$ retrieves, schedules, and executes atomic skills corresponding to $\{\ell^a_x, \dots, \ell^a_z\}$, enabling zero-shot generalization.

\subsection{Physical Interaction-based Task Decomposition and Skill Learning}\label{method:skills}
To construct modular and reusable atomic skills, \algabrvname{} proposes a novel task decomposition strategy inspired by human hand-object interactions and prior work~\cite{chen2024gravmad,saito2023structured}. This decomposition is based on the physical interactions of the robotic gripper, as described in Sec.~\ref{method:formula}.
For the original demonstrations, $\mathcal{T}^o$, used to train the multi-task IL model $\mathcal{M}$, \algabrvname{} decomposes tasks based on full physical interactions $p^{\texttt{full}}$. For instance, a demonstration for the instruction \textit{``put item in a closed drawer without closing the drawer''} can be divided into two atomic tasks: \textit{``open drawer''} and \textit{``place item into open drawer''}, each aligned with a full gripper interaction.
After decomposition, DeCo reformats the atomic demonstrations for skill learning. Each atomic task $T^a_i$ is paired with a language instruction $\ell^a_i$, forming the instruction library $\mathcal{L}^{a}$. Demonstrations are processed using a keyframe discovery method~\cite{james2022q} that identifies keyframes based on gripper state transitions or near-zero joint velocities. Each demonstration concludes with a full physical interaction $p^{\texttt{full}}$, and the end-effector pose in the final keyframe is marked as the goal pose defined in the robot base frame.
Optionally, demonstrations may include temporal data (e.g., time steps) to support task progression modeling. Finally, $\mathcal{M} + \mathrm{DeCo}$ is trained with these physically consistent atomic datasets, enabling it to effectively acquire multiple atomic skills.
The objective is to learn a language-conditioned policy $\pi^a_{\theta^*}$ that maps observation-instruction pairs to actions:
$\theta^* = \arg\min_{\theta} \, \mathbb{E}_{i, (o, a)} \left[ \mathcal{L}_{\text{MT-IL}}\left( \pi^a_\theta(o, \ell^a_i), a \right) \right],$
where $\pi^a_{\theta^*}(o, \ell^a_i) = \mathcal{M}_{\theta^*}(o, \ell^a_i)$.
\textbf{Unless otherwise stated, all training datasets of atomic tasks and experimental results of $\mathcal{M} + \mathrm{DeCo}$ presented in the main paper are based on $p^{\texttt{full}}$.}
To explore the suitable granularity of physical interaction, we also implement a DeCo variant based on $p^{\texttt{half}}$. Ablation study results are discussed in Sec.~\ref{sec:experiments:ablation}.

\subsection{VLM-Guided Planning and Skill Scheduling}\label{method:scheduling}

After a multi-task imitation learning (IL) model $\mathcal{M}$ has mastered multiple atomic skills, DeCo enables it to generalize to novel long-horizon tasks through skill composition. Given a novel yet compositional task $T^{\text{new}}$, $\mathcal{M}+\text{DeCo}$ first invokes the vision--language model (VLM) GPT-4o~\cite{hurst2024gpt} with the task instructions, observed RGB images, and a pre-built atomic instruction library. Leveraging the VLM’s reasoning and planning capabilities, the system retrieves relevant atomic instructions and produces an ordered skill sequence to accomplish the task.
The low-level execution of each atomic skill is carried out entirely by the IL policy $\mathcal{M}$, while DeCo provides only goal-pose supervision and skill-level scheduling. To connect consecutive skills, DeCo does not rely on a conventional global motion planner; instead, it employs a short-horizon, spatially constrained motion-bridging module. This module generates collision-aware transitional motions between the previous skill’s goal pose and the next skill’s start pose, while keeping the trajectory close to the demonstrated state distribution, thereby mitigating policy distribution shift.
During execution, the system continuously monitors the robot pose and checks it against the goal pose of the current skill. A successful match indicates completion of the atomic skill—corresponding to a full gripper interaction cycle—and triggers the execution of the next skill; otherwise, the current skill continues until the goal pose is reached. This pose-based closed-loop monitoring mechanism ensures reliable skill termination without modifying the underlying policy.
Additional implementation details, including VLM prompts, are provided in the supplementary materials on our project website.

\subsection{Spatially-Aware Skill Chaining}\label{method:chaining}

Although $\mathcal{M} + \text{DeCo}$ can semantically combine atomic skills to accomplish long-horizon tasks via VLM-guided planning and scheduling (see Sec.~\ref{method:scheduling}), challenges remain in executing them sequentially. A key issue is achieving smooth transitions between atomic skills in 3D space. Each skill learned by $\mathcal{M}$ has distinct start and goal poses, often causing large spatial discontinuities between successive skills. Traditional motion planners, while avoiding collisions, lack semantic-spatial awareness of policy compatibility. They may select geometrically feasible transitions that lie in low-confidence regions of the next policy, leading to failures.  
To address this, DeCo introduces a spatially-aware skill chaining module that enables seamless transitions without modifying pose distributions. Once the current skill completes—i.e., the robot reaches the goal pose—the system schedules the next instruction and predicts its start pose. The current goal pose, predicted next start pose, and scene point cloud are passed to a spatially-aware cost map module adapted from Voxposer~\cite{huang2023voxposer}. This module generates collision-free chaining poses bridging the two skills, as shown in the third part of \autoref{fig:overview}. The robot then performs RRT-based~\cite{karaman2011anytime} motion planning over these poses to ensure safe transitions. Operating during skill handoff in $\mathcal{M} + \text{DeCo}$, this module enables smooth composition and reliable execution of long-horizon 3D manipulation.

\section{\texttt{DeCoBench}: Benchmark Overview}
\label{app:decobench}

\begin{figure}[t!]
    \vspace{5pt}
    \centering
    \includegraphics[width=\linewidth]{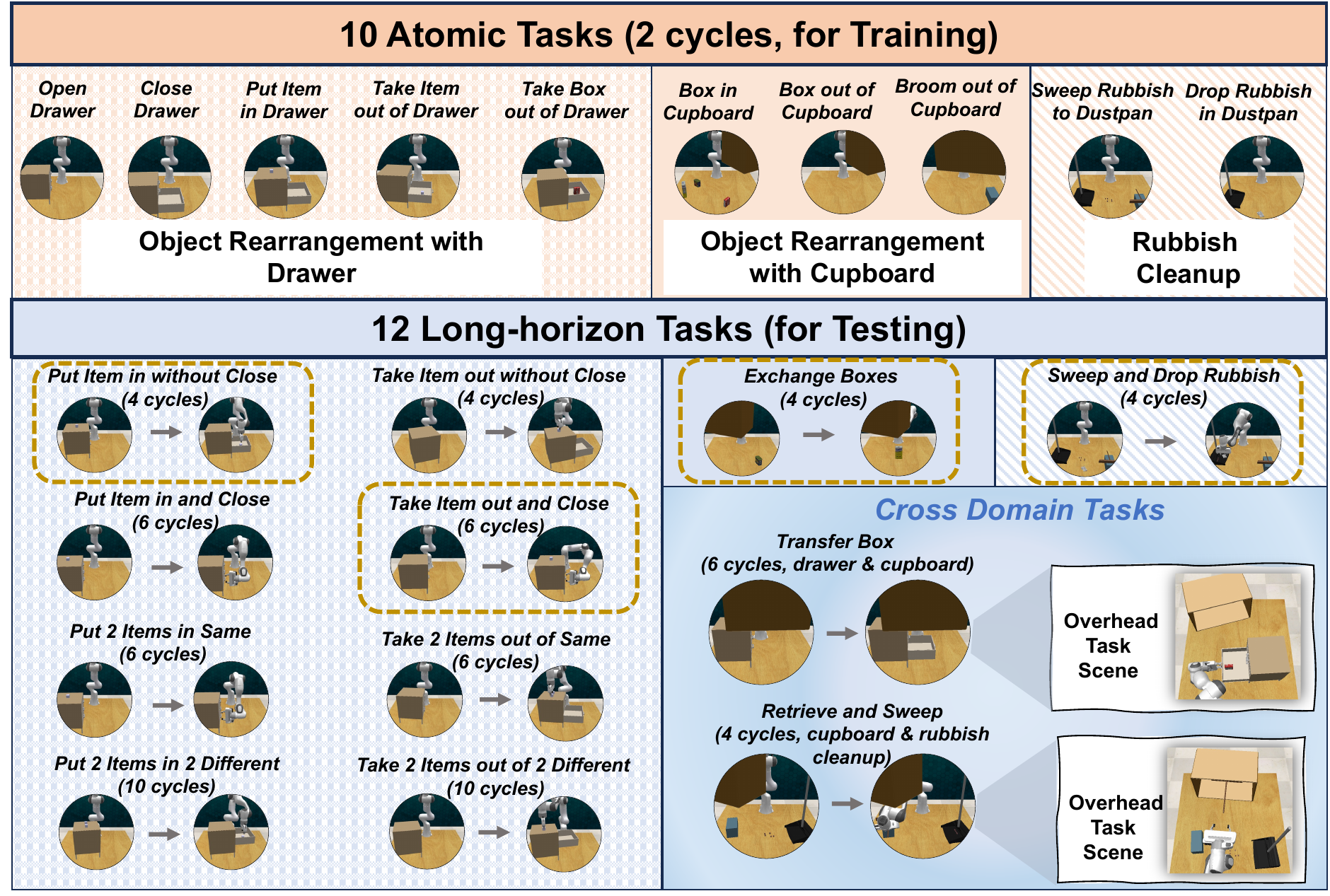}
    \caption{
    % \textcolor{black}{\small \textbf{\texttt{DeCoBench} Overview}. We decompose 4 source tasks (yellow dashed boxes) into 10 reusable atomic skills via physical interaction analysis (Sec.~\ref{method:skills}). These skills are then recomposed into 12 novel long-horizon tasks—covering within- and cross-domain settings—to benchmark the zero-shot compositional generalization of multi-task IL models.}
    \small \textbf{\texttt{DeCoBench} Overview}. We decompose 4 source tasks (yellow dashed boxes) into 10 reusable atomic skills via physical interaction analysis (Sec.~\ref{method:skills}). These skills are recomposed into 12 novel long-horizon tasks—spanning within- and cross-domain settings—to benchmark zero-shot compositional generalization.
    }
    \label{fig:decobench}
    \vspace{-1mm}
\end{figure}

\begin{figure}[t]
    \centering
    \includegraphics[width=\linewidth]{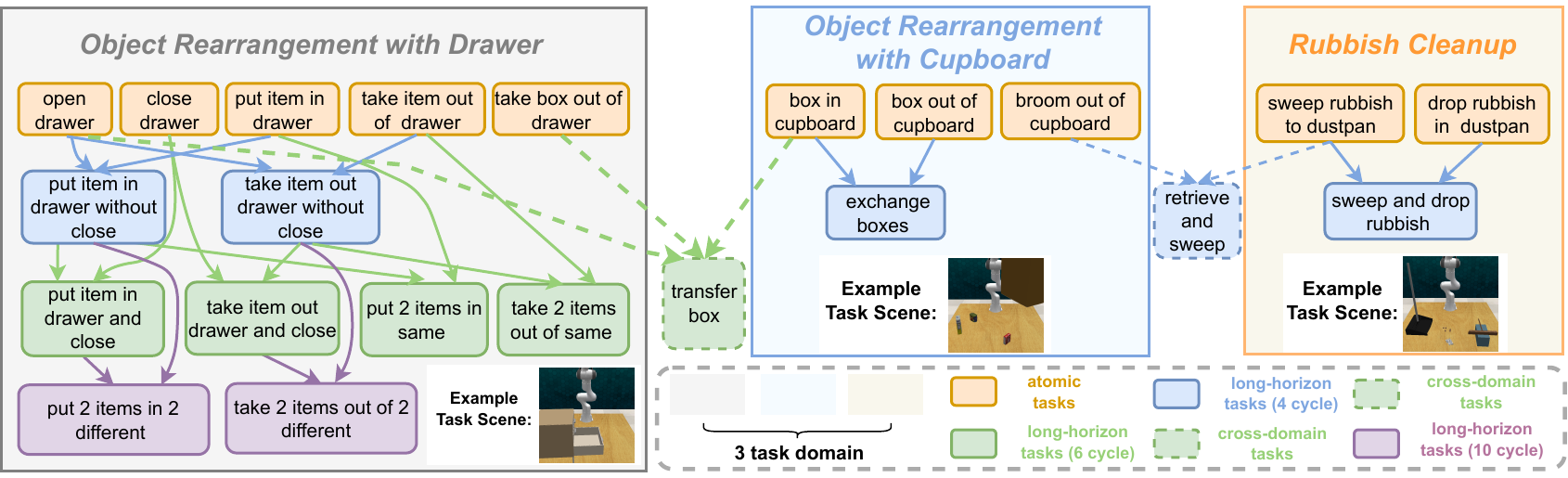}
    \caption{
        \textcolor{black}{\small Compositional hierarchy in \texttt{DeCoBench}, illustrating how long-horizon tasks are composed of atomic tasks.} 
    } 
    \label{fig:decobench_relations}
    % \vspace{-8mm}
\end{figure}

\texttt{DeCoBench} is presented as a benchmark for evaluating zero-shot generalization in novel compositional long-horizon 3D manipulation tasks, as shown in \autoref{fig:decobench}. While \texttt{DeCoBench} is heavily inspired by RLBench~\cite{james2020rlbench}, it is specifically designed to focus on the compositionality of tasks and their ability to generalize across different task compositions. \texttt{DeCoBench} covers three domains: \textbf{Object Rearrangement with Drawer}, \textbf{Object Rearrangement with Cupboard}, and \textbf{Rubbish Cleanup}.
In the \textbf{Object Rearrangement with Drawer} domain, original tasks—\textit{put item in drawer without close} and \textit{take item out of drawer and close}—are decomposed into four atomic tasks based on the gripper’s interaction cycle: \textit{open drawer}, \textit{close drawer}, \textit{put item in drawer}, and \textit{take item out of drawer}. A variant atomic task, \textit{take box out of drawer}, is created via object substitution. This domain includes two 4-cycle tasks, five 6-cycle tasks, and two 10-cycle tasks.
In the \textbf{Object Rearrangement with Cupboard} domain, \textit{exchange boxes} is decomposed into \textit{box in cupboard} and \textit{box out of cupboard}, with \textit{take broom out of cupboard} introduced as an additional task. The \textit{exchange boxes} task serves as a 4-cycle compositional long-horizon task.
In the \textbf{Rubbish Cleanup} domain, \textit{sweep and drop rubbish} is decomposed into \textit{sweep rubbish to dustpan} and \textit{drop rubbish in dustpan}, with the original task serving as a 4-cycle compositional task.
The benchmark further incorporates two cross-domain compositional tasks: \textit{transfer box} (6 cycles) across Drawer and Cupboard, and \textit{retrieve and sweep} (4 cycles) across Cupboard and Cleanup. \autoref{fig:decobench_relations} illustrates the compositional relationships between atomic and long-horizon tasks.

\section{Experiments}
\label{sec:result}
\begin{table*}[t]
\centering
% \vspace{5pt}
\caption{\small \textbf{Test Performance on 10 atomic tasks in \texttt{DeCoBench}.} Evaluations on 10 atomic tasks are conducted using 3 seeds, with 20 test episodes per task, utilizing the final checkpoints from training on 10 atomic tasks.}
\renewcommand{\arraystretch}{0.6}
\resizebox{0.9\textwidth}{!}{
\begin{tabular}{lcccccccccccc} 
\toprule
\rowcolor[HTML]{CBCEFB}
         & Avg.                & Open            & Close             & Put in Opened             & Take Out of           & Box Out of             & Box in         & Box Out    & Broom Out   & Sweep to          & Rubbish in\\
\rowcolor[HTML]{CBCEFB}
Models  & Success $\uparrow$   & Drawer          & Drawer            & Drawer                 & Opened Drawer            & Opened Drawer          & Cupboard       & Cupboard   & Cupboard    & Dustpan           & Dustpan\\
\midrule
RVT-2~\cite{goyal2024rvt2} & 91.83 
& 98.33 {$_{\pm 2.36}$} 
& 96.67 {$_{\pm 2.36}$} 
& 100.00 {$_{\pm 0.00}$}
& 100.00  {$_{\pm 0.00}$}
& 100.00  {$_{\pm 0.00}$}
& 35.00 {$_{\pm 4.08}$} 
& 98.33 {$_{\pm 2.36}$}  
& 98.33 {$_{\pm 2.36}$} 
& 91.67 {$_{\pm 6.24}$} 
& 100.00 {$_{\pm 0.00}$}  \\
RVT-2+DeCo & 86.80 
& 98.33 {$_{\pm 2.36}$} 
& 100.00  {$_{\pm 0.00}$}
& 88.33 {$_{\pm 6.24}$}  
& 100.00  {$_{\pm 0.00}$}
& 100.00  {$_{\pm 0.00}$}
& 48.33 {$_{\pm 2.36}$} 
& 85.00 {$_{\pm 7.07}$} 
& 65.00  {$_{\pm 0.00}$}
& 83.00 {$_{\pm 6.24}$} 
& 100.00 {$_{\pm 0.00}$} \\
\midrule
3DDA~\cite{ke20243d} & 98.00 
& 98.33 {$_{\pm 2.36}$} 
& 100.00  {$_{\pm 0.00}$}
& 100.00  {$_{\pm 0.00}$}
& 100.00  {$_{\pm 0.00}$}
& 98.33 {$_{\pm 2.36}$} 
& 91.67 {$_{\pm 4.71}$} 
& 98.33 {$_{\pm 2.36}$} 
& 96.67 {$_{\pm 2.36}$} 
& 98.33 {$_{\pm 2.36}$} 
& 100.00 {$_{\pm 0.00}$} \\
3DDA+DeCo & 96.00 
& 95.00  {$_{\pm 0.00}$}
& 100.00  {$_{\pm 0.00}$}
& 100.00  {$_{\pm 0.00}$}
& 100.00  {$_{\pm 0.00}$}
& 100.00  {$_{\pm 0.00}$}
& 88.33 {$_{\pm 2.36}$} 
& 100.00  {$_{\pm 0.00}$}
& 98.33 {$_{\pm 2.36}$}
& 78.33 {$_{\pm 2.36}$}
& 100.00 {$_{\pm 0.00}$} \\
\midrule
ARP~\cite{zhang2025autoregressive} & 94.67 
& 100.00 {$_{\pm 0.00}$}
& 95.00 {$_{\pm 0.00}$}
& 100.00 {$_{\pm 0.00}$}
& 100.00 {$_{\pm 0.00}$}
& 100.00 {$_{\pm 0.00}$}
& 65.00 {$_{\pm 7.07}$} 
& 95.00 {$_{\pm 0.00}$}
& 100.00 {$_{\pm 0.00}$}
& 93.33 {$_{\pm 2.36}$} 
& 98.33 {$_{\pm 2.36}$}  \\
ARP+DeCo & 91.67 
& 100.00 {$_{\pm 0.00}$}
& 100.00 {$_{\pm 0.00}$} 
& 100.00  {$_{\pm 0.00}$}
& 100.00 {$_{\pm 0.00}$} 
& 100.00  {$_{\pm 0.00}$}
& 30.00 {$_{\pm 7.07}$} 
& 95.00 {$_{\pm 0.00}$}
& 96.67 {$_{\pm 2.36}$} 
& 96.67 {$_{\pm 2.36}$} 
& 98.33 {$_{\pm 2.36}$} \\
\bottomrule
\end{tabular}
}
% \vspace{-3mm}
\label{tab:app_atomic}
\end{table*}

\begin{table*}[t]
\centering
\caption{\small \textbf{Generalization Performance on \texttt{DeCoBench} Long-horizon tasks.} 
% Final checkpoints are tested across 3 seeds with 20 episodes per task.
Above is a visualization illustrating how DeCo enables zero-shot generalization on two long-horizon tasks from \texttt{DeCoBench}.
}
\vspace{-2mm}
\includegraphics[width=0.85\textwidth]{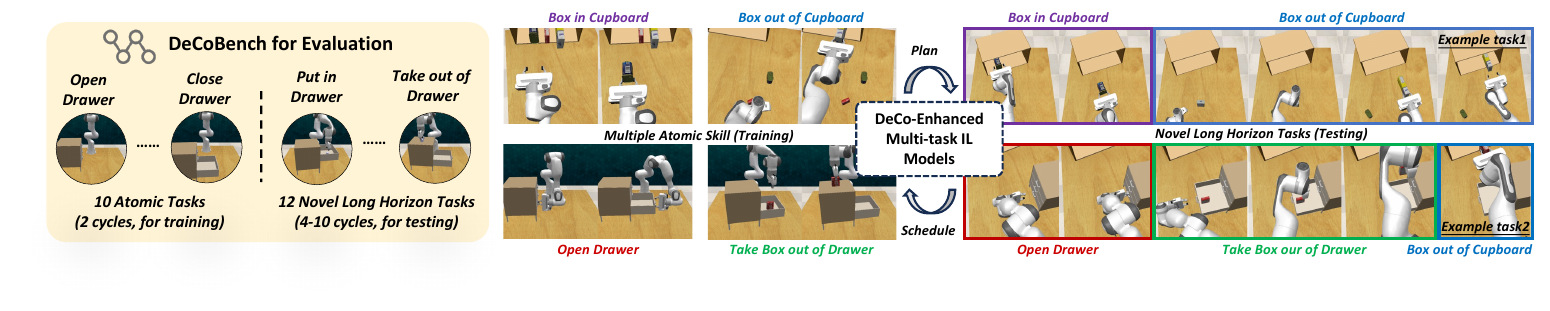} 
\vspace{-5mm} % 调整图片和表格之间的垂直间距

\renewcommand{\arraystretch}{0.6}
\resizebox{0.85\textwidth}{!}{
\begin{tabular}{lcccccccccccccc} 
\toprule
\rowcolor[HTML]{CBCEFB}
& Avg. & Put in & Put in & Take out & Take out & Put Two & Take Two & Put Two & Take Two & Exchange & Sweep & Transfer & Retrieve \\
\rowcolor[HTML]{CBCEFB}
Models & Success $\uparrow$ & w/o Close & and Close & w/o Close & and Close & in Same & out of Same & in Diff & out of Diff & Boxes &and Drop &  Box & and Sweep \\
\midrule
RVT2~\cite{goyal2024rvt2} & 0.00 & 0.00 & 0.00 & 0.00 & 0.00 & 0.00 & 0.00 & 0.00 & 0.00 & 0.00 & 0.00 & 0.00 & 0.00 \\
\rowcolor[HTML]{EFEFEF}
RVT2 + DeCo & \underline{\textbf{66.67}} \textcolor{blue}{(66.67\% $\uparrow$)} & 
\underline{98.33} {$_{\pm 2.36}$} & 
\underline{98.33}  {$_{\pm 2.36}$} & 
\underline{93.33} {$_{\pm 6.24}$} & 
\underline{96.67} {$_{\pm 4.71}$} & 
\underline{93.33} {$_{\pm 6.24}$} &
\underline{71.67} {$_{\pm 12.47}$} & 
\underline{85.00} {$_{\pm 7.07}$} & 
\underline{61.67} {$_{\pm 17.00}$} & 
\underline{11.67} {$_{\pm 6.24}$} & 
\underline{80.00} {$_{\pm 4.08}$} & 
0.00  & 
\underline{10.00} {$_{\pm 4.08}$} \\
\midrule
3DDA~\cite{ke20243d} & 0.00 & 0.00 & 0.00 & 0.00 & 0.00 & 0.00 & 0.00 & 0.00 & 0.00 & 0.00 & 0.00 & 0.00 & 0.00 \\
\rowcolor[HTML]{EFEFEF}
3DDA + DeCo & \underline{\textbf{21.53}} \textcolor{blue}{(21.53\% $\uparrow$)} & 
0.00 & 
0.00 & \underline{83.33} {$_{\pm 9.43}$} & 
\underline{68.33} {$_{\pm 4.71}$} & 0.00 & 
0.00 & 
0.00 & 0.00 & 
\underline{95.00} {$_{\pm 4.08}$}& 0.00 & \underline{11.67} {$_{\pm 2.36}$} & 0.00 \\
\midrule
ARP~\cite{zhang2025autoregressive} & 0.14 & 0.00 & 0.00 & 0.00 & 0.00 & 0.00 & 0.00 & 0.00 & 0.00 & 0.00 & \underline{1.67} {$_{\pm 2.36}$} & 0.00 & 0.00 \\
\rowcolor[HTML]{EFEFEF}
ARP + DeCo & \underline{\textbf{58.06}} \textcolor{blue}{(57.92\% $\uparrow$)} & 
\underline{96.67} {$_{\pm 4.71}$} & 
\underline{95.00} {$_{\pm 0.00}$} & 
\underline{96.67} {$_{\pm 2.36}$} & 
\underline{96.67} {$_{\pm 2.36}$}& 
\underline{98.33} {$_{\pm 2.36}$}& 
\underline{71.67} {$_{\pm 20.14}$}& 
\underline{76.67} {$_{\pm 4.71}$}& 
0.00 & 
\underline{63.33} {$_{\pm 2.36}$}& 
0.00 & 
0.00 & 
\underline{1.67} {$_{\pm 2.36}$} \\
\midrule

% Newly Added Rows
% \rowcolor[HTML]{FFFFFF}
% \textcolor{black}{$\pi_{0}$~\cite{black2410pi0}} &
% \textcolor{black}{0.00} &
% \textcolor{black}{0.00} &
% \textcolor{black}{0.00} &
% \textcolor{black}{0.00} &
% \textcolor{black}{0.00} &
% \textcolor{black}{0.00} &
% \textcolor{black}{0.00} &
% \textcolor{black}{0.00} &
% \textcolor{black}{0.00} &
% \textcolor{black}{0.00} &
% \textcolor{black}{0.00} &
% \textcolor{black}{0.00} &
% \textcolor{black}{0.00} \\

% \rowcolor[HTML]{FFFFFF}
% \textcolor{black}{$\pi_{0}$ + DeCo} &
% \textcolor{black}{\underline{\textbf{38.33}}} \textcolor{blue}{(38.33\% $\uparrow$)} &
% \textcolor{black}{75.00 $_{\pm 7.07}$} &
% \textcolor{black}{73.33 $_{\pm 8.08}$} &
% \textcolor{black}{52.67 $_{\pm 6.24}$} &
% \textcolor{black}{61.67 $_{\pm 4.71}$} &
% \textcolor{black}{65.00 $_{\pm 6.12}$} &
% \textcolor{black}{47.00 $_{\pm 10.71}$} &
% \textcolor{black}{59.33 $_{\pm 5.46}$} &
% \textcolor{black}{42.67 $_{\pm 9.43}$} &
% \textcolor{black}{9.33 $_{\pm 4.08}$} &
% \textcolor{black}{56.67 $_{\pm 4.71}$} &
% \textcolor{black}{0.00 $_{\pm 0.00}$} &
% \textcolor{black}{6.67 $_{\pm 3.24}$} \\

\bottomrule
\end{tabular}
}
\vspace{-4mm}
\label{tab:main_decobench}
\end{table*}

DeCo is evaluated in both simulated and real-world environments. Specifically, the following research questions are addressed:
1) To what extent does DeCo enhance the generalization of multi-task IL models on long-horizon 3D manipulation tasks (Sec.~\ref{sec:experiments:generalization})?
2) Can the framework achieve robust generalization in real-world long-horizon manipulation tasks (Sec.~\ref{sec:experiments:real})?
3) How do heuristic settings in DeCo influence its generalization performance (Sec.~\ref{sec:experiments:ablation})?

\subsection{Experimental Setup}
\label{sec:experiments:setup}

\textbf{Baseline Multi-task IL Models.}
DeCo is applied to three representative multi-task IL models—RVT-2~\cite{goyal2024rvt2}, 3DDA~\cite{ke20243d}, and ARP~\cite{zhang2025autoregressive}—to validate its model-agnostic design and demonstrate its generalization benefits.
RVT-2 is a multi-view robotic transformer using a coarse-to-fine strategy on point clouds to predict the next-best action heatmap.
3DDA combines 3D scene representations with a diffusion-based policy for manipulation.
ARP employs a Chunking Causal Transformer~\cite{zhang2025autoregressive} to autoregressively generate action sequences for manipulation tasks.

\textbf{Simulation Setup.}
Simulation experiments are conducted on the proposed \texttt{DeCoBench} benchmark. Demonstrations are generated using scripted policies, with goal poses defined in the robot base frame.
Observations are collected from four RGB-D cameras at the front, left shoulder, right shoulder, and wrist.
RVT-2 and ARP use $128{\times}128$ image inputs, while 3DDA uses $256{\times}256$, following original settings.
Each baseline and its DeCo-enhanced variant are trained on atomic tasks (50 demonstrations per task) and evaluated on compositional tasks (20 test demonstrations per task).
All policies are evaluated with three random seeds.
For fair comparison, original training configurations are used: batch size 24 for RVT-2, 48 for ARP, and 8 for 3DDA.
All models are trained on 8 NVIDIA GeForce RTX 4090 GPUs.

\begin{figure}
    % \vspace{3pt}
    \centering
    \includegraphics[width=\linewidth]{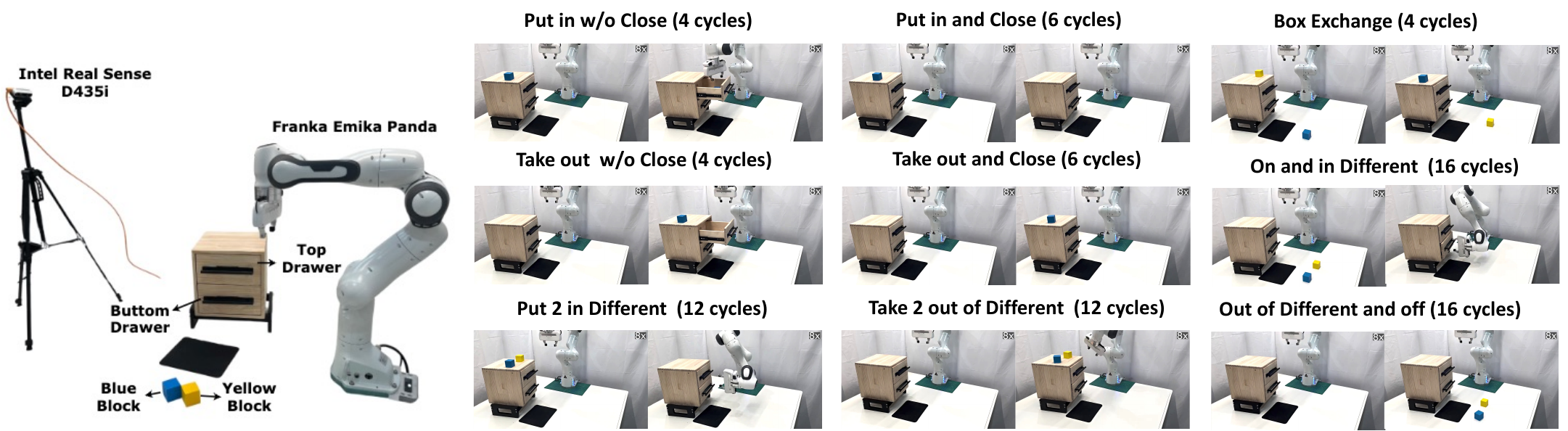}
    \caption{\small \textbf{Left:} Real-Robot Setup. \textbf{Right:} 9 Real-World Compositional Long-Horizon Tasks.}
    \label{fig:real_&_long}
    % \vspace{-8mm}
\end{figure}

\textbf{Real-robot Setup.}
DeCo is validated on a Franka Emika Panda robot equipped with an exocentric Intel RealSense D435i camera, as shown in \autoref{fig:real_&_long} (left).
RVT-2 and RVT-2+DeCo are compared on an object rearrangement task involving a drawer.
Training uses 6 atomic tasks (16 variations) collected via kinesthetic teaching ($\approx$5 mins per task), while testing covers 9 long-horizon tasks (30 variations) for zero-shot generalization.
As shown in \autoref{fig:real_&_long} (right), the test set includes 3 tasks with 4 cycles, 2 with 6, 2 with 12, and 2 with 16 cycles.
Each task is executed 10 times with randomized initial object placements to compute average success rates.

\subsection{Generalization Performance on \texttt{DeCoBench}}
\label{sec:experiments:generalization}

\autoref{tab:app_atomic} shows the performance of various models on 10 atomic tasks in \texttt{DeCoBench}. While 3DDA achieves the highest overall success rate, ARP and RVT-2 also demonstrate competitive performance. However, when DeCo is integrated into these models, some tasks show a decrease in performance. This can be attributed to the instability introduced by the VLM during atomic task planning, which negatively affects the accuracy of instruction-following in multi-task IL models. This drop primarily stems from VLM visual-semantic grounding errors: 1) state estimation hallucinations (e.g., falsely detecting occlusions) trigger unnecessary prerequisite skills (over-planning), disrupting the execution flow; and 2) instruction distribution shifts, where VLM-generated instructions differ from training data, causing the IL policy to estimate suboptimal goal poses.
Notably, baseline models rely on ground-truth instructions they encountered during training, which enables them to perform better on these tasks.
In contrast, \autoref{tab:main_decobench} highlights DeCo’s effectiveness in novel task generalization. It presents the performance of RVT-2, 3DDA, and ARP on 12 long-horizon tasks after being trained on 10 atomic tasks, alongside their DeCo-enhanced counterparts. While the baseline models excel at atomic tasks (as shown in \autoref{tab:app_atomic}), they struggle significantly with long-horizon tasks, failing to generalize atomic skills to more complex scenarios.
This failure stems from two fundamental limitations: the inability to reason about unseen temporal dependencies and perceptual aliasing in circular loops. Without explicit task decomposition, baselines cannot infer necessary pre-conditions (e.g., opening before placing) from high-level instructions. Moreover, in tasks with repeated states, these reactive policies struggle to differentiate identical observations across different stages.
DeCo, however, substantially boosts their performance, improving the success rates from 0.00\% to \textbf{66.67\%} for RVT-2+DeCo, from 0.00\% to \textbf{21.53\%} for 3DDA+DeCo, from 0.14\% to \textbf{58.06\%} for ARP+DeCo.
% , and from 0.00\% to \textbf{38.33\%} for $\pi_0$+DeCo.
These results underscore the power of DeCo’s model-agnostic design and its capacity for zero-shot generalization, efficiently scheduling and composing learned atomic skills to perform well in new, long-horizon tasks.
It is important to emphasize that the upper bound of DeCo's enhancement on model generalization is limited by: (1) the inherent capability of the base IL model, and (2) the task reasoning ability of the VLM. For example, although 3DDA+DeCo and ARP+DeCo perform well on training atomic tasks, they fail on certain compositional long-horizon tasks such as \textit{Sweep and Drop} (sweep rubbish + drop rubbish) and \textit{Retrieve and Sweep} (broom out of cupboard + sweep rubbish). While DeCo can accurately plan and schedule the corresponding atomic skills, both 3DDA and ARP face challenges in visual processing within these unseen compositional contexts.

\begin{table}[h]
\vspace{-3mm}
 \captionof{table}{\small \textbf{Impact of Atomic Task Design.}}
  \centering
  \footnotesize
    \setlength\tabcolsep{4pt}
    \renewcommand{\arraystretch}{0.03}
    \resizebox{0.8\linewidth}{!}{
    \begin{tabular}{@{}lccc@{}}
        \toprule
        % \cmidrule(lr){3-4}
        \textbf{Task} & \textbf{RVT-2+DeCo} & \textbf{RVT-2 (6 Long training)}  \\
        \midrule
        6 Novel& \textbf{83.89\%} \textcolor{blue}{(53.89\% $\uparrow$)} &  30.00\%  \\
         % 6 L-H  & 49.45\% & \textbf{74.72\%}  \\
        \midrule
        12 All  & \textbf{66.67\%} \textcolor{blue}{(14.31\% $\uparrow$)} &  52.36\% \\
        \bottomrule
    \end{tabular}}
    % \vspace{-1mm}
    \label{tab:limitation}
    % \vspace{-3mm}
\end{table}

To further assess the impact of atomic task design, we train RVT-2 on 6 long-horizon tasks (6 Long) from \texttt{DeCoBench}, comprising 4 original IL tasks and 2 cross-domain tasks.  
We then evaluate the model on the remaining 6 novel long-horizon tasks not seen during training (6 Novel), as well as on all 12 long-horizon tasks (12 All). 
As shown in \autoref{tab:limitation}, compared to RVT-2 trained directly on the 6 long-horizon tasks \textcolor{black}{, denoted as \textbf{RVT-2 (6 Long training)}}, the DeCo-based variant (RVT-2 + DeCo) achieves substantially better zero-shot generalization on the unseen 6 Novel tasks, improving the success rate by \textbf{53.89\%}.  
Even when evaluated across all 12 tasks, including those seen by RVT-2 (6 Long training), the DeCo-based model still yields an overall improvement of \textbf{14.31\%}.  
These results indicate that DeCo’s atomic task training set more effectively supports skill acquisition in multi-task IL models and enhances generalization to novel compositional tasks. 

\begin{table*}[t!]
\centering
\vspace{5pt}
\caption{\small \textbf{Real-world results.} 
Each entry represents the successful trials out of 10. 
Above is a visualization example showing how DeCo performs zero-shot generalization in one of the longest-horizon tasks \textit{On and in Different} (16 cycles). 
% Above the table are illustrations of the 9 novel long-horizon (N-L-H) tasks, with the number of cycles ranging from 4 to 16.
}
% 插入图片
\includegraphics[width=0.75\textwidth]{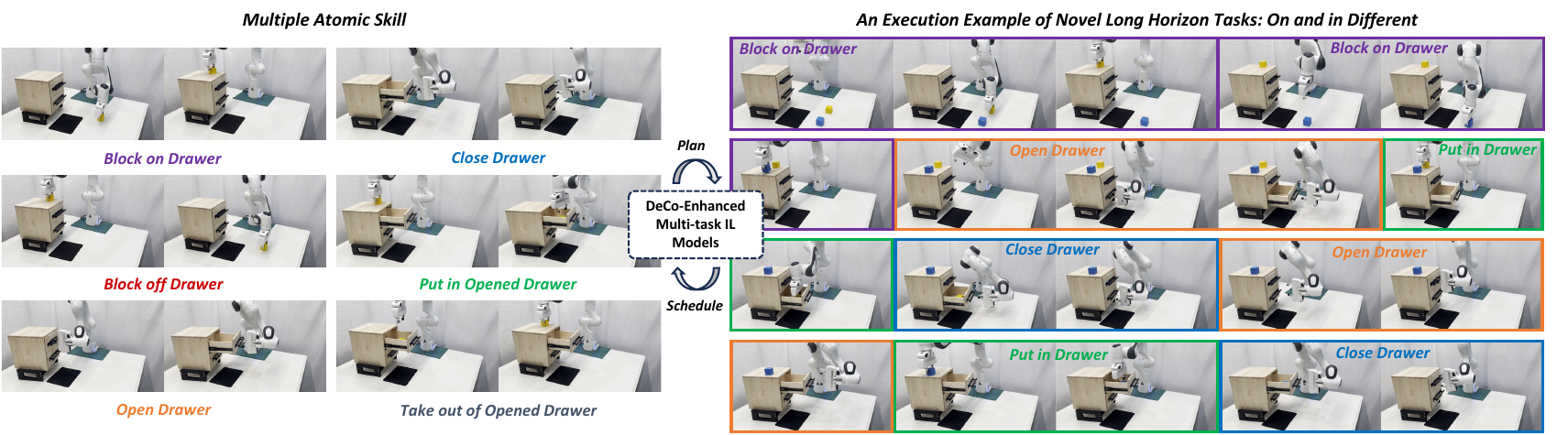} 
\vspace{-0.1mm} % 调整图片和表格之间的垂直间距
\small
\renewcommand{\arraystretch}{0.5} % 更紧凑行距
\captionsetup{skip=4pt} % 减小标题与表格的垂直间距
\resizebox{0.75\textwidth}{!}{\begin{tabular}{lcc|lcc}
\toprule
\textbf{Atomic Tasks} & \textbf{RVT2} & \textbf{RVT2+DeCo} & 
\textbf{N-L-H Tasks} & \textbf{RVT2} & \textbf{RVT2+DeCo} \\
\midrule
Open Drawer                  & 8/10  & 8/10  &
Put in w/o Close         & 0/10  & \textbf{7/10} \\
Close Drawer                 & 9/10  & 9/10  & 
Put in and Close             & 0/10  & \textbf{7/10}  \\
Put in Opened Drawer         & 9/10 & 8/10 & 
Take out w/o Close       & 0/10  & \textbf{6/10}  \\
Take out of Opened Drawer    & 9/10 & 8/10 & 
Take out and Close           & 0/10  & \textbf{5/10}  \\
Block on Drawer              & 10/10 & 10/10 & 
Put 2 in Different    & 0/10  & \textbf{4/10}  \\
Block off Drawer             & 10/10 & 10/10 & 
Take 2 out of Different    & 0/10  & \textbf{3/10}  \\ 

                             &       &       & 
                             Block Exchange               & 0/10  & \textbf{9/10}  \\
                             &       &       & 
                             On and in Different          & 0/10  & \textbf{3/10}  \\
                             &       &       & 
                             Out of Different and off     & 0/10  & \textbf{1/10}  \\
\midrule
\textbf{Avg. SR on Atomic Tasks} & 91.67\% & 88.33\% & \textbf{Avg. SR on N-L-H Tasks} & 0\% & \textbf{53.33\%} \\
\bottomrule
\end{tabular}}
\label{tab:real}
\vspace{-1mm}
\end{table*}

\begin{figure*}[t!]
  \centering
  \begin{subfigure}[b]{0.28\textwidth}
    \centering
    \includegraphics[width=\textwidth]{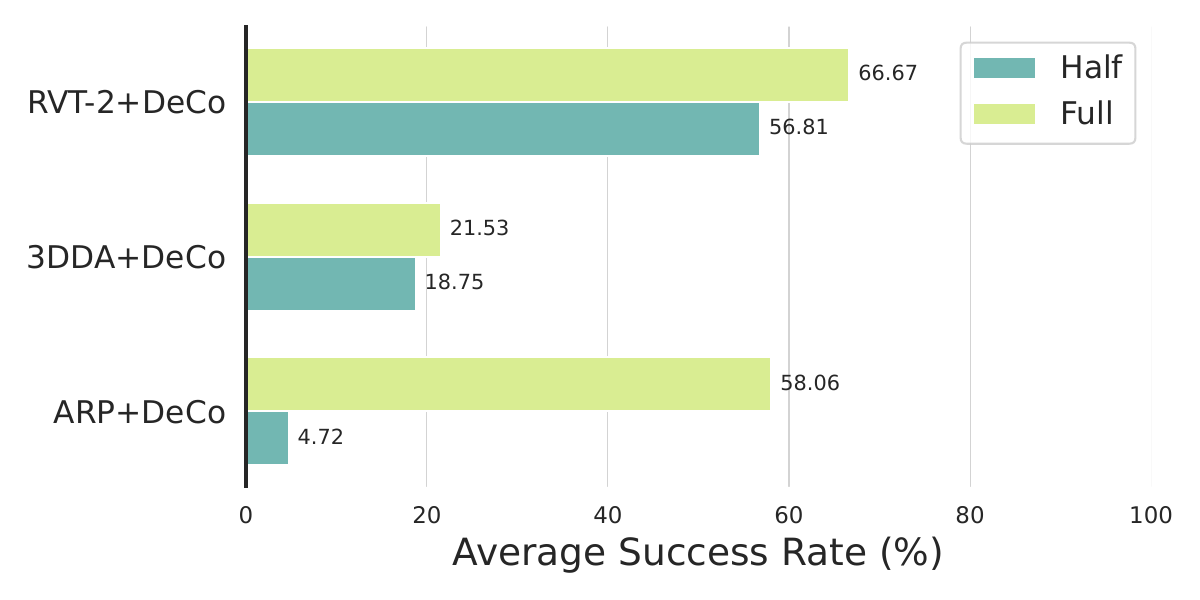}
    \caption{\small{Half vs. Full Interaction}}
    \label{exp:half_vs_full}
  \end{subfigure}
  \hfill
  \begin{subfigure}[b]{0.28\textwidth}
    \centering
    \includegraphics[width=\textwidth]{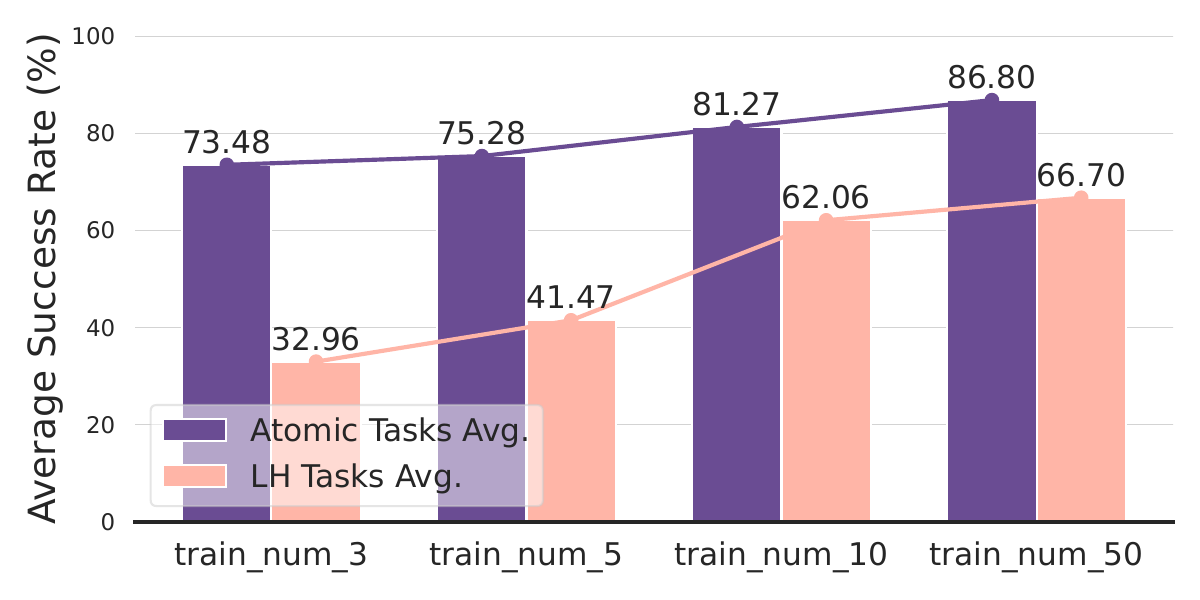}
    \caption{\small{Effect of Training Demo Num}}
    \label{exp:demo_num}
  \end{subfigure}
  \hfill
  \begin{subfigure}[b]{0.28\textwidth}
    \centering
    \includegraphics[width=\textwidth]{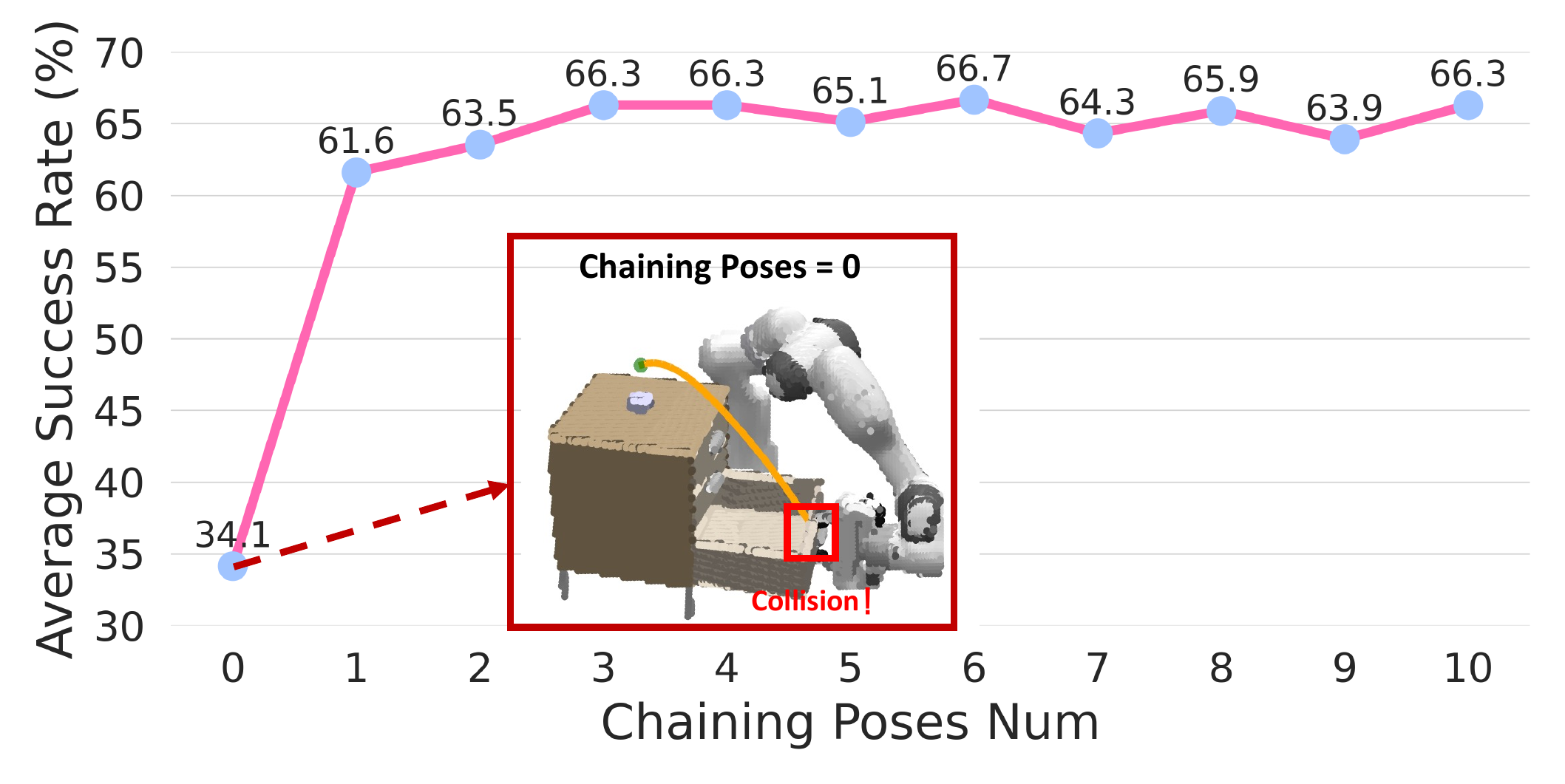}
    \caption{\small{Effect of Chaining Poses Num}}
    \label{exp:cp_num}
  \end{subfigure}
  \vspace{-1.1mm}
  \caption{\small \textbf{Ablation study of heuristic settings in DeCo.} (b) and (c) are based on the RVT-2+DeCo model.}
  \label{fig:ablation}
  \vspace{-4mm}
\end{figure*}

\subsection{Real-robot Evaluations}
\label{sec:experiments:real}

Extensive experiments are conducted on a real-world robotic platform to validate the practical effectiveness of the DeCo framework, assessing its ability to generalize learned skills to novel instructions and scenarios. \autoref{tab:real} compares the zero-shot success rates of RVT-2 and RVT-2+DeCo, both trained on 6 atomic tasks, and evaluated on 9 novel long-horizon (N-L-H) tasks unseen during training. These N-L-H tasks require reasoning over task instructions and composing atomic skills to achieve desired outcomes.
The results show that RVT-2 struggles with generalization, achieving a success rate of 0\%. In contrast, \textbf{RVT-2+DeCo achieves an average success rate of 53.33\% on the N-L-H tasks,} highlighting the practical effectiveness and generalization of DeCo in real-world settings.

To analyze system robustness, we introduce human intervention during real-world task execution and evaluate two perturbation scenarios. When the target object is displaced, the end-effector fails to reach the goal pose; since DeCo determines task completion via pose matching, the monitoring module detects the mismatch and halts execution, exposing the lack of a re-planning mechanism. In contrast, under object replacement or minor drawer perturbations, DeCo correctly identifies task completion and schedules subsequent skills as long as the goal pose is reached.
These results show that DeCo is robust to perturbations preserving goal poses but cannot recover from significant object displacement. Related experiment videos are available on the project website.

\subsection{Ablation Studies}
\label{sec:experiments:ablation}

\autoref{fig:ablation} summarizes three ablation studies on DeCo’s heuristic settings. 
\autoref{exp:half_vs_full} compares the generalization performance of three models using DeCo under half and full interaction settings. The results show that while DeCo enhances generalization in both cases, different models exhibit varying sensitivity to the granularity of physical interactions. Notably, \textbf{full interaction sequences (open → closed → open) significantly improve DeCo's compositional generalization capability}, with ARP+DeCo being most affected in the half interaction setting, while RVT-2+DeCo and 3DDA+DeCo show relatively stable performance.
We believe that full interaction decomposition generates clear and consistent subtasks, providing stronger learning signals and enhancing generalization. In contrast, half interaction decomposition often results in overly fine-grained fragments, increasing task interference and hindering policy learning. While finer decomposition allows for greater flexibility, it also expands the task space and raises the risk of inconsistent subtask combinations.
Additionally, full interaction decomposition aligns better with language instructions, as each decomposition usually corresponds to a complete command. This alignment is crucial for effectively bridging perception, action, and language in DeCo. 

The number of atomic task demonstrations for RVT-2+DeCo is varied, reporting the average success rates over 10 atomic tasks and 12 long-horizon tasks. As shown in \autoref{exp:demo_num}, more demonstrations improve the performance of the IL model. \textbf{Provided the RVT-2 model learns atomic skills to a sufficient degree, DeCo effectively composes these skills to achieve zero-shot generalization on long-horizon tasks}. Moreover, improved atomic task performance directly correlates with enhanced generalization in DeCo.
\autoref{exp:cp_num} shows that disabling the spatially-aware skill chaining module (Chaining Poses Num = 0) causes a significant performance drop. A failure case of the \textit{Put in and Close} task shows a collision with the drawer due to poor transition planning, preventing successful grasping. \textbf{In contrast, enabling the spatially-aware skill chaining module consistently improves task success, regardless of the number of poses}. In DeCo, the chaining poses number is heuristically set to 6.
\section{Conclusion and Discussion}
\label{sec:conclusion}

This paper proposes \textbf{DeCo}, a model-agnostic framework enabling multi-task IL models to zero-shot generalize to novel compositional long-horizon 3D manipulation tasks. DeCo decomposes IL demonstrations into modular atomic tasks through physical interaction analysis, leverages VLMs for task planning, and uses a spatially-aware chaining module for collision-free transitions. Extensive evaluations in both simulation and real-world settings show that DeCo significantly improves the generalization of multi-task IL models.

Limitations arise from system dependencies. Our analysis identifies two failure modes: atomic task failures due to VLM visual-semantic grounding errors (e.g., over-planning), and long-horizon task failures from the base IL model’s limited visual robustness in novel contexts. Additionally, the current decomposition is limited to claw-like end-effectors.
Future work includes exploring tactile modalities for dexterous control and non-prehensile manipulation, as well as incorporating closed-loop re-planning and parameterized primitives to enhance robustness and scalability.

%===============================================================================

\bibliographystyle{IEEEtran}
\bibliography{main}  % .bib

\end{document}